\documentclass[letterpaper]{article} 
\usepackage{aaai2027}  
\usepackage[hyphens]{url}  
\usepackage{graphicx} 
\usepackage{natbib}  
\usepackage{caption} 
\usepackage{algorithm}
\usepackage{algorithmic}
\usepackage{amsmath}
\usepackage{amssymb}
\usepackage{amsfonts}
\usepackage{booktabs}
\usepackage{multirow}
\usepackage{amsmath}

\usepackage{newfloat}
\usepackage{listings}
\DeclareCaptionStyle{ruled}{labelfont=normalfont,labelsep=colon,strut=off} 
\floatstyle{ruled}
\newfloat{listing}{tb}{lst}{}
\floatname{listing}{Listing}

\usepackage{booktabs}

\nocopyright

\title{SHSP: Structure-Aware Hierarchical Solution Prediction for Mixed-Integer Linear Programming}
\author{
    Zherong Zhang\textsuperscript{\rm 1,\rm 2}\equalcontrib,
    Guanlin Li\textsuperscript{\rm 1,\rm 2}\equalcontrib,
    Chengrui Gao\textsuperscript{\rm 1,\rm 2},
    Haopu Shang\textsuperscript{\rm 1,\rm 2},
    Ke Xue\textsuperscript{\rm 1,\rm 2},
    Jixiang Lu\textsuperscript{\rm 3},
    Weiyong Yang\textsuperscript{\rm 3},
    Chao Qian\textsuperscript{\rm 1,\rm 2}\corresponding
}
\affiliations{
    \textsuperscript{\rm 1}State Key Laboratory of Novel Software Technology, Nanjing University\\
    \textsuperscript{\rm 2}School of Artificial Intelligence, Nanjing University\\
    \textsuperscript{\rm 3}State Key Laboratory of Technology and Equipment for Defense Against Power System Operational Risks, Nari Technology Co., Ltd.
}

\begin{document}

\maketitle

\begin{abstract}
Mixed-Integer Linear Programming (MILP) is a fundamental optimization paradigm in combinatorial optimization and has been widely applied across real-world domains. Due to its NP-hard nature, obtaining optimal solutions for large-scale or highly constrained MILP instances remains computationally prohibitive. Learning-based solution prediction has therefore emerged as a promising approach to provide high-quality variable assignment for solver acceleration. However, existing methods typically adopt a one-shot prediction paradigm that predicts the marginal probabilities of all variables simultaneously. As a result, the conditional dependencies among variables are only implicitly captured through message passing, with the burden of modeling the combinatorial structure falling entirely on the representational capacity of graph neural networks. To address this limitation, we propose the \textbf{S}tructure-Aware \textbf{H}ierarchical \textbf{S}olution \textbf{P}rediction (SHSP) framework that replaces the parallel marginal decoding of one-shot methods with a novel hierarchical conditional decoding mechanism. Specifically, SHSP constructs a variable coupling graph from the constraint structure, decodes variables sequentially along a hierarchy of increasing coupling strength, and conditions each hierarchy on previously predicted assignments. To mitigate error accumulation during the decoding process, SHSP further incorporates a confidence-aware mask-and-repair mechanism to identify and correct unreliable intermediate predictions. We integrate SHSP with multiple learning-guided search methods, and evaluate it on four standard MILP benchmarks. Experimental results demonstrate that SHSP significantly outperforms existing one-shot prediction baselines, achieving a 54\% average reduction in solution gap. Code is available at: \url{https://github.com/lamda-bbo/SHSP}.
\end{abstract}

\section{Introduction}
Mixed-Integer Linear Programming (MILP) is one of the most widely used modeling paradigms in combinatorial optimization and operations research, with broad applications in various optimization domains~\cite{ma2019accelerating,li2024b}. Owing to the NP-hardness of MILP~\cite{karp1972reducibility}, decades of research efforts have been devoted to developing powerful solvers such as SCIP~\cite{achterberg2009scip} and Gurobi~\cite{gurobi2021}. Equipped with advanced techniques including cutting planes, primal heuristics, and presolving, these solvers are capable of efficiently handling a wide range of practical instances. However, solving large-scale and highly constrained instances remains computationally expensive, making the improvement of MILP solving efficiency a fundamental research challenge.

In recent years, machine learning has emerged as a promising paradigm for accelerating MILP solving~\cite{li2024optverse}, and existing studies generally fall into two categories. The first category learns heuristic policies for key components of branch-and-bound (B\&B) solvers, such as branching strategy~\cite{khalil2016learning,gasse2019exact,hou2026llm4branch} and cut selection~\cite{tang2020reinforcement,huang2022learning,paulus2022learning,miracle2026}, typically via imitation learning or reinforcement learning. The second category employs machine learning models such as graph neural networks (GNNs) to directly predict high-quality solutions for MILP instances~\cite{ding2020accelerating,zeng2024effective,geng2025differentiable}. Research in this category advances along two complementary aspects. The first aspect concerns how predictions are exploited to reduce the original problem~\cite{nair2020solving,han2023gnn,liu2025apollo}. The second aspect focuses on improving the prediction capacity~\cite{pu2026coco,wang2026milpnet}. A detailed discussion of related work is provided in Appendix~\ref{sec:Related Works}. Together, these efforts have established learning-based solution prediction as an effective paradigm for accelerating MILP solving.

Despite the remarkable performance of existing solution prediction methods, most of them follow a one-shot prediction paradigm~\cite{han2023gnn,liu2025apollo,pu2026coco}, where the marginal probabilities of all decision variables are predicted simultaneously in a single forward pass. However, variables in a MILP are strongly interdependent through shared constraints. The optimal assignment of one variable often hinges on the combinatorial structure formed by the others, to the extent that exactly predicting the complete optimal assignments would amount to solving the MILP itself. Under the one-shot paradigm, such dependencies are captured only implicitly by the GNN encoder, while the decoding stage treats all variables as independent. This mismatch can yield mutually conflicting assignments that mislead, rather than accelerate, the downstream solver.

To address these limitations, we propose \textbf{\underline{S}tructure-Aware \underline{H}ierarchical \underline{S}olution \underline{P}rediction for MILP~(SHSP)}, a novel learning framework that explicitly models variable dependencies through hierarchical partitioning and prediction, rather than relying on a one-shot predictor to recover them implicitly. Specifically, SHSP first constructs a weighted variable coupling graph that quantifies pairwise dependencies based on expected constraint violation and coefficient importance. 
Guided by the coupling scores, variables are predicted in a hierarchical manner: weakly coupled variables are resolved first, while strongly coupled ones are subsequently decoded conditioned on the resulting partial assignments. To further enhance robustness, we introduce a confidence-based mask-and-repair mechanism that mitigates error accumulation during multi-step inference. Finally, the predicted assignments are exploited through a structure-aware fixing strategy that prioritizes strongly coupled variables, whose fixed values propagate through shared constraints and tighten the feasible domains of many others, thereby effectively reducing the search space.

As a superior predictor, SHSP serves as a drop-in replacement for the vanilla GNN-based one-shot predictor and can be seamlessly integrated into diverse learning-guided acceleration frameworks~\cite{nair2020solving,han2023gnn}. To validate the effectiveness of SHSP, we conduct extensive experiments on four widely used MILP benchmarks. Experimental results demonstrate that SHSP consistently outperforms the one-shot prediction baselines across all benchmarks, reducing the average absolute primal gap by up to 100.0\%, 38.1\%, 94.3\%, and 42.1\%, respectively. Notably, a variant of SHSP even surpasses the solution quality of full-budget Gurobi on combinatorial auction instances within a substantially smaller time budget.

\section{Preliminaries}
\subsection{Mixed-Integer Linear Programming}
Mixed-integer linear programming (MILP) seeks to optimize a linear objective function over a feasible region defined by linear constraints, where a subset of decision variables take integer values. Formally, an MILP instance can be expressed as
\begin{equation*}
\label{eq:milp}
\begin{aligned}
\min_{x} \quad
    & c^{\top}x \\
\text{s.t.} \quad
    & Ax \le b, \\
    & l \le x \le u, \\
    & x \in \mathbb{Z}^{p}
      \times \mathbb{R}^{n-p},
\end{aligned}
\end{equation*}
where \(x\) denotes the \(n\)-dimensional decision vector comprising \(p\) integer variables and \(n-p\) continuous variables. The vector \(c \in \mathbb{R}^{n}\) specifies the objective coefficients, \(A \in \mathbb{R}^{m \times n}\) is the constraint coefficient matrix, and \(b \in \mathbb{R}^{m}\) is the corresponding right-hand-side vector. The vectors \(l \in (\mathbb{R} \cup \{-\infty\})^{n}\) and \(u \in (\mathbb{R} \cup \{+\infty\})^{n}\) impose the lower and upper bounds on the variables, respectively. Following prior works~\cite{han2023gnn,liu2025apollo,pu2026coco}, we focus on MILP instances whose integer variables are binary, as general integer variables can be reduced to binary ones via well-established preprocessing techniques~\cite{nair2020solving}.

An MILP instance can be naturally encoded as a weighted bipartite graph $G=(W\cup V, E)$~\cite{gasse2019exact}, where $W$ and $V$ denote the sets of constraint nodes and variable nodes, respectively. An edge $(w, v) \in E$ connects a constraint node $w \in W$ and a variable node $v \in V$ if and only if the corresponding variable appears with a nonzero coefficient in the constraint. The node features typically encompass objective coefficients, variable bounds, and variable types on the variable side, as well as right-hand-side values and constraint senses on the constraint side, while the edge features carry the corresponding constraint coefficients. More details on the graph modeling are provided in Appendix~\ref{sec:Graph Representation Details}.

\subsection{Predict-and-Search}

Predict-and-Search (PaS)~\cite{han2023gnn} is a two-stage framework that integrates neural solution prediction with mathematical optimization for accelerating MILP solving. Given an MILP instance $I$, PaS first learns the underlying solution distribution from a collection of optimal or near-optimal solutions. Specifically, the solution distribution is approximated as $q(x|I)=\frac{\exp(-E(x,I))}{\sum_{x'\in S}\exp(-E(x',I))}$, where $S$ denotes the collected solution set and the energy function is defined as $E(x,I)=c^\top x$ if $x$ is feasible and $+\infty$ otherwise. Guided by this distribution, PaS trains a graph neural network
$p_\theta$ to estimate the marginal probability of each binary
variable. Under the variable-wise independence assumption, the
joint solution distribution is factorized as
$p_\theta(x\mid I)=\prod_{i=1}^{n}p_\theta(x_i\mid I)$,
thereby reducing the learning task to predicting
per-variable marginals. The GNN is trained by minimizing a
weighted binary cross-entropy loss over the collected
solutions, where each solution is weighted according to
$q(x\mid I)$:
\begin{equation*}
\mathcal{L}
=
-
\sum_{s=1}^{S}
q(x^{(s)}\mid I)
\sum_{i=1}^{n}
\ell(\hat p_i,y_{s,i})
\end{equation*}
where
$\hat{p}_i=p_\theta(x_i=1\mid I)$
denotes the predicted probability of variable $x_i$,
$y_{s,i}\in\{0,1\}$ denotes the value of variable $x_i$
in solution $x^{(s)}$, and
$\ell(\hat p_i,y_{s,i})
=
-
y_{s,i}\log\hat p_i
-
(1-y_{s,i})\log(1-\hat p_i)$
denotes the binary cross-entropy loss.
Based on the predicted marginal probabilities, PaS selects $k_1$ variables with the highest marginal probabilities and fixes them to one, while selecting $k_0$ variables with the lowest marginal probabilities and fixes them to zero. The resulting partial solution is denoted as $\hat{x}^{[P]}$, where $P$ represents the index set of fixed variables.

Rather than hard-fixing these variables as in neural diving~\cite{nair2020solving}, PaS formulates a trust-region optimization problem that constrains the feasible region to the neighborhood of the predicted partial solution. Formally, the trust region is defined as $B_P(\hat{x}^{[P]},\Delta)=\{x^{[P]}\in\mathbb{R}^{n}\mid \|\hat{x}^{[P]}-x^{[P]}\|_1\leq\Delta\}$, where $\Delta$ denotes the neighborhood radius. The resulting trust-region problem is formulated as

\begin{equation*}
\begin{aligned}
\min_{x}\quad & c^\top x\\
\text{s.t.}\quad
& Ax\leq b,\quad l\leq x\leq u,\\
& x^{[P]}\in B_P(\hat{x}^{[P]},\Delta),\quad x\in\mathbb{Z}^{p}\times\mathbb{R}^{n-p}.
\end{aligned}
\end{equation*}

By coupling neural prediction with trust-region-based local optimization, PaS effectively combines the efficiency of learning-based methods and the reliability of mathematical programming solvers. Notably, when $\Delta=0$, PaS degenerates to the hard-fixing scheme of neural diving.

\section{Methodology}

\begin{figure*}[t]
    \centering
    \includegraphics[width=\textwidth]{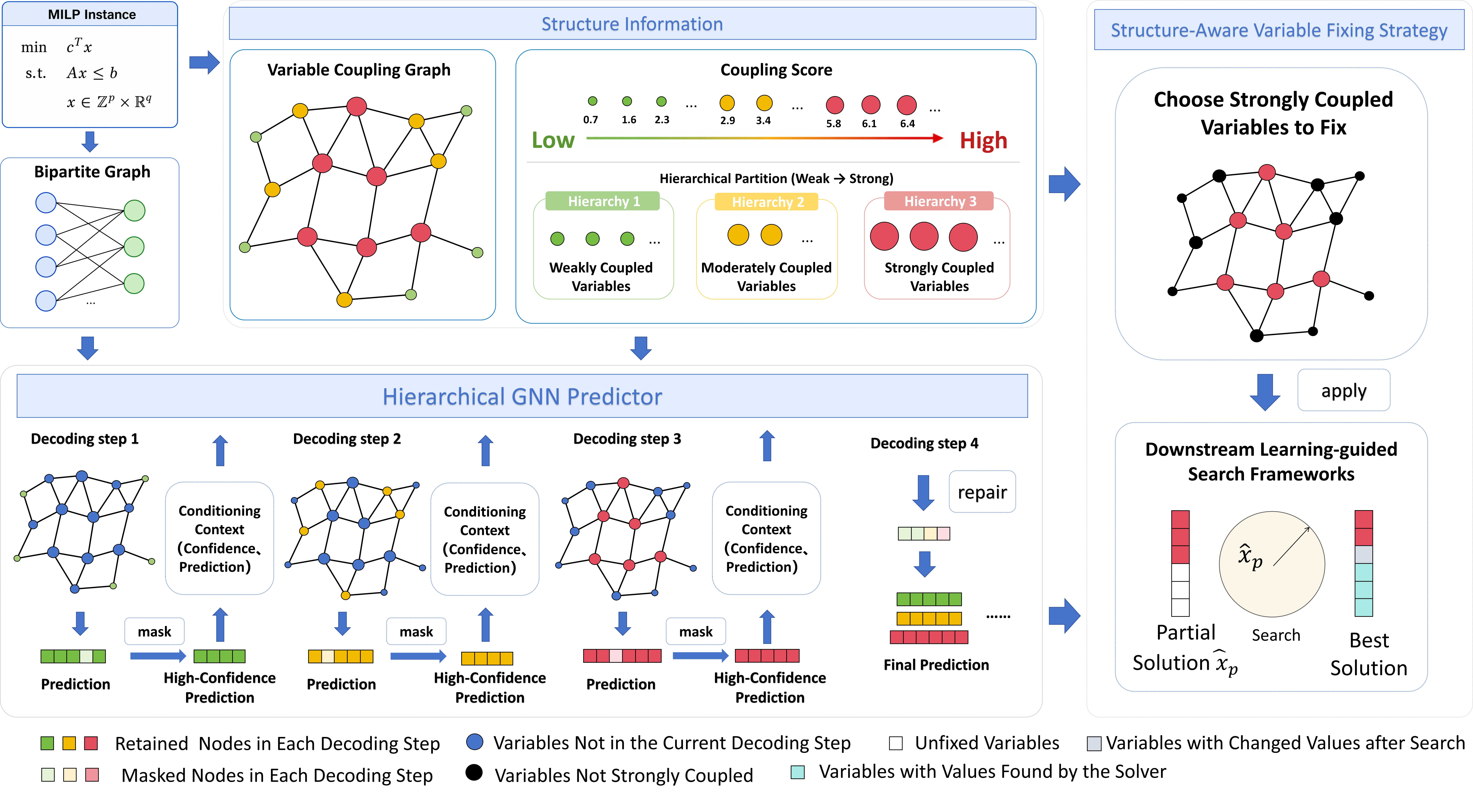}
    \caption{Overview of the proposed method. Given an original MILP instance, we first
construct a variable coupling graph to capture the structural dependencies among decision variables. Based on variable coupling graph, variables are ranked according to their coupling score and divided into multiple hierarchies. Our SHSP sequentially predicts variables following the weak-to-strong hierarchy, with a mask-and-repair mechanism to mitigate error accumulation. Finally, the predicted variable assignments are combined with a structure-aware variable fixing strategy and applied to downstream learning-guided search frameworks such as Predict-and-Search~\cite{han2023gnn}.}
    \label{fig:framework}
\end{figure*} 

In this section, we present \textbf{\underline{S}tructure-Aware \underline{H}ierarchical \underline{S}olution \underline{P}rediction for MILP~(SHSP)}. In contrast to existing methods that predict the assignments of all decision variables simultaneously, SHSP explicitly exploits variable coupling information and decodes variable assignments sequentially following a hierarchy of increasing coupling strength, where the prediction at each decoding step is conditioned on the partial assignments of lower hierarchies obtained in preceding steps. The overall framework is illustrated in Figure~\ref{fig:framework}. In the remainder of this section, we first introduce the variable coupling graph, then describe the structure-aware hierarchical predictor, and finally present the corresponding variable fixing strategy.

\subsection{Variable Coupling Graph}
\label{sec:coupling_graph}

Decision variables are often tightly coupled through shared constraints, and such structural dependencies play a critical role in obtaining high-quality solutions. To explicitly capture these dependencies, we construct a variable coupling graph, which serves as the basis for estimating coupling scores and for determining hierarchical partitioning.

\paragraph{Graph Modeling.} We model the coupling structure among variables as an undirected weighted graph, where nodes correspond to variables and edges capture their co-occurrence in constraints. Given linear constraints, we consider all variables when estimating coupling strengths, but retain only those that co-occur with at least one binary variable in some constraints, i.e., those directly connected to binary variables. This makes the coupling graph considerably sparser without severely compromising the prediction accuracy of binary variables. An edge is created between two retained variables if they appear together in at least one constraint, with its weight quantifying the coupling strength implied by the shared constraints, as detailed below.

\paragraph{Edge Weights Computation.}

For each retained variable pair, we assign an edge weight that reflects the coupling strength implied by the constraints. The weight is determined by two complementary heuristic factors. The first factor, termed the \emph{expected violation}, measures how strongly a variable pair is restricted by a constraint. Specifically, we treat the two variables as random variables, where binary variables follow independent uniform distributions over $\{0,1\}$ and continuous variables follow uniform distributions over their bounds. We then compute the expected violation of the constraint induced by the pair alone. A larger expected violation indicates that the pair is more tightly restricted by the constraint and thus more strongly coupled. The second factor, termed the \emph{coefficient importance}, characterizes the relative contribution of a variable pair within a constraint. It is computed as the sum of the magnitudes of the two corresponding coefficients, each scaled by the range of its associated variable. To eliminate scale discrepancies across constraints, both factors are normalized by their respective average values over all variable pairs within the same constraint. Note that the above two factors are both computed over all variables, including purely continuous ones.

The local coupling weight of a pair in constraint $c$ is then given by the product of the two normalized factors, and the final edge weight $W_{ij}$ between variables $z_i$ and $z_j$ is obtained by summing the local weights over all constraints containing both variables. The detailed computation of edge weights is provided in Appendix~\ref{sec:Variable Coupling Graph Details}. The resulting weighted graph captures the overall structural coupling among variables and serves as the basis for computing the variable coupling score and determining the hierarchical prediction order in structure-aware hierarchical prediction.

\subsection{Structure-Aware Hierarchical Prediction}
\label{sec:hierarchical}

The above weighted variable coupling graph provides a quantitative measure of the structural dependency among variables. Building upon this, we design a structure-aware hierarchical prediction framework that decodes variables sequentially from weakly coupled to strongly coupled ones. This framework conditions each decoding step on the assignments fixed in preceding hierarchies, so that strongly coupled variables, whose values are the most sensitive to those of others, are predicted with explicit structural context rather than in isolation as in previous one-shot methods.

\paragraph{Coupling Score-based Hierarchical Partitioning.}
For each binary variable $z_i$, we define its variable coupling score (VCS) as the sum of the weights of its incident edges in the coupling graph, which quantifies the overall strength of its structural dependency on the remaining variables. All variables are sorted in ascending order of VCS and evenly partitioned into $K$ hierarchies $\{H_1, H_2, \ldots, H_K\}$, such that $H_1$ contains the most weakly coupled variables and $H_K$ the most strongly coupled ones. The decoding process then follows this weak-to-strong order. Weakly coupled variables are largely insensitive to the assignments of other variables and can thus be predicted reliably at early stages. Their fixed values in turn serve as conditioning information that anchors the prediction of strongly coupled variables in later hierarchies.

\paragraph{Hierarchical Decoding with Mask-and-Repair.}

The model decodes the $K$ hierarchies sequentially, and each step is conditioned on the tentative assignments accumulated from all preceding hierarchies. To realize this conditioning, the input feature of every variable node is augmented with a decoding-state triplet $(m_i, \hat{x}_i, c_i)$, where $m_i \in \{0,1\}$ indicates whether variable $i$ should be predicted in the current step, $\hat{x}_i$ denotes its tentative binary assignment, and $c_i$ its associated confidence. For variables that will be decoded in future steps, the triplet is set to a zero placeholder. In addition, to make the network aware of its position in the decoding process, the current step index is encoded via positional encoding~\cite{vaswani2017attention}, and the resulting step embedding is concatenated with the initial node embeddings and passed through a fusion layer to produce the step-aware node representations. The collection of these state features constitutes the conditioning context $\mathbf{h}_{<k}$, and the prediction of the $k$-th hierarchy is formulated as
\begin{equation*}
\mathbf{p}_{H_k}
=
f_{\theta}\!\left(G, H_k, \mathbf{h}_{<k}\right),
\end{equation*}
where $G$ denotes the bipartite graph and $f_{\theta}$ the graph neural network. For each variable $i \in H_k$, the network outputs a probability $p_i$, from which the binary assignment $\hat{x}_i = \mathbb{I}(p_i > 0.5)$ and the confidence $c_i = 2\lvert p_i - 0.5 \rvert$ are derived.

Since the conditioning context is constructed from model predictions rather than ground-truth assignments, wrong early decisions may propagate and accumulate across hierarchies. To mitigate such error accumulation, we propose a mask-and-repair mechanism. Before being incorporated into the conditioning context, each newly predicted variable is stochastically masked with a probability $p_{\mathrm{mask}}(i) \propto 1 - c_i$ that decreases with its confidence. Consequently, high-confidence predictions are likely to be retained as conditioning information, while uncertain predictions tend to be masked, reverting to the undecided placeholder state and being collected into a repair set $\mathcal{R}$. After all $K$ hierarchies have been decoded, a final repair step decodes the variables in $\mathcal{R}$ conditioned on the full context of retained assignments, i.e., \(\mathbf{p}_{\mathcal{R}}
=
f_{\theta}\!\left(G, \mathcal{R},
\mathbf{h}_{\mathcal{V}\setminus\mathcal{R}}\right),\)
and the repaired predictions replace the masked ones to yield the final solution. This iterative masking-and-prediction paradigm shares a similar spirit with the recent work Apollo-MILP~\cite{liu2025apollo}. However, our method differs in two key aspects: (1) it explicitly exploits the variable coupling structure to establish a hierarchy that guides the decoding process; (2) it constitutes an inherently new prediction paradigm that operates entirely within the prediction stage, rather than following Apollo-MILP's alternation between prediction and solver invocation.

\paragraph{Training Objective.}
The framework is trained by minimizing a weighted prediction loss aggregated over all hierarchies and the repair step. Given $S$ supervised solutions with objective values $\{o_s\}_{s=1}^{S}$, each solution is assigned a weight $w_s = \exp(-o_s/\tau) \,/\, \sum_{j=1}^{S}\exp(-o_j/\tau)$, where higher-quality solutions contribute more to the supervision and $\tau$ controls the weighting smoothness. The overall training objective is defined as
\begin{equation*}
\begin{split}
\mathcal{L}
=&\;
\frac{1}{K}
\sum_{k=1}^{K}
\sum_{s=1}^{S} w_s
\sum_{i\in H_k}
\ell\!\left(p_i^{(k)}, y_{s,i}\right) \\
&+
\sum_{s=1}^{S} w_s
\sum_{i\in\mathcal{R}}
\ell\!\left(p_i^{(\mathrm{rep})}, y_{s,i}\right),
\end{split}
\end{equation*}
where $p_i^{(k)}$ and $p_i^{(\mathrm{rep})}$ denote the predicted probabilities of variable $i$ in the $k$-th hierarchy and the repair step, respectively, $y_{s,i}$ denotes the label of variable $i$ in solution $s$, and $\ell(\cdot,\cdot)$ is the binary cross-entropy loss. Note that when computing the $k$-th prediction, the preceding $k-1$ states are decoded by the model itself, which aligns with the inference procedure and thus mitigates the distributional shift between training and inference. However, fully relying on model-decoded states can hinder convergence, as early predictions are highly inaccurate. We therefore incorporate a \textit{teacher forcing} strategy that employs ground-truth assignments as conditioning information in the early stage of training, and the proportion of teacher forcing is progressively decayed as training proceeds~(See Appendix~\ref{subsec:Training Details}). 

\subsection{Structure-Aware Variable Fixing Strategy}
\label{sec:fixing}
Existing variable fixing strategies typically select variables solely according to their prediction confidences. However, fixing variables with similar confidence levels may have substantially different structural influences on the reduced MILP problem, so treating them uniformly overlooks the valuable structural information. 

Motivated by this observation, we propose a structure-aware variable fixing strategy that jointly considers the variable coupling structure and the prediction confidences. Specifically, we first identify variables whose variable coupling scores exceed a prescribed threshold as high-coupling variables. Fixing such variables tends to tighten the feasible domains of many others through shared constraints, and thus effectively reduces the search space. The identified variables are subsequently screened according to their prediction confidences, where variables with $p_i \ge \theta_1$ are fixed to one, and those with $p_i \le \theta_0$ are fixed to zero. This two-stage screening guarantees that only strongly coupled variables with highly reliable predictions are eligible for fixing.

To align with existing methods and ensure a fair comparison, we further cap the number of fixed variables: the variables selected above are ranked by their prediction probabilities, and at most $k_1$ and $k_0$ of them are retained for fixing to one and to zero, respectively. Note that $k_1$ and $k_0$ serve only as upper bounds. If the thresholds yield fewer variables, no additional ones are fixed to reach these bounds. In contrast to confidence-only fixing, the proposed strategy prioritizes variables that are both structurally influential and confidently predicted, thereby enabling downstream optimization frameworks to explicitly exploit structural information for more effective search-space reduction.

\section{Experiments}
\subsection{Experimental Setup}
\paragraph{Datasets}
We evaluate our framework on four widely adopted MILP benchmarks in this field: Combinatorial Auctions (CA)~\cite{leytonbrown2000towards}, Workload Appointment (WA)~\cite{gasse2022ml4co}, Item Placement (IP)~\cite{gasse2022ml4co}, and Set Covering (SC)~\cite{balas1980set}. Following existing studies~\cite{gasse2019exact,han2023gnn,liu2025apollo}, we generate SC and CA instances using similar procedures, while the IP and WA benchmarks are obtained from two real-world problems used in NeurIPS ML4CO 2021 competition
~\cite{gasse2022ml4co}. For each benchmark problem, we use 240 instances for training, 60 instances for validation, and 100 instances for testing. Additional details are provided in Appendix~\ref{sec:Benchmark Details}.

\paragraph{Baselines}
We compare our method with both traditional MILP solvers and
learning-based solution prediction approaches. For traditional optimization
methods, we compare our method with Gurobi~\cite{gurobi2021} and SCIP~\cite{achterberg2009scip}.
For learning-based approaches, we consider three representative methods: Neural Diving (ND)~\cite{nair2020solving}, Predict-and-Search
(PaS)~\cite{han2023gnn}, and Apollo-MILP~\cite{liu2025apollo}. 

\paragraph{Metrics.}
We compare the best
objective values (OBJ) achieved by different methods within a 1000-second
time limit on each test instance. Following the evaluation protocol in
Predict-and-Search~\cite{han2023gnn}, we use the best solution obtained by a single-threaded
Gurobi solver with a 3600-second time limit as the best-known solution (BKS),
which serves as an approximation of the optimal value. The absolute primal
gap is then calculated as
Gap$_{abs}=|$OBJ$-$BKS$|$,
measuring the distance between the obtained solution and the BKS. It is worth noting that our method achieves better solutions than the Gurobi solver with a 3600-second time limit on certain benchmark problems. In such cases, the objective values found by our method are used to update the BKS.

\begin{table*}[t]
\centering

\renewcommand{\arraystretch}{1.15}
\setlength{\tabcolsep}{4pt}

\small
\begin{tabular}{lcccccccc}
\toprule
\multirow{2}{*}{Method}
& \multicolumn{2}{c}{CA $\uparrow$~(BKS:98489.24)}
& \multicolumn{2}{c}{WA $\downarrow$~(BKS:706.86)}
& \multicolumn{2}{c}{IP $\downarrow$~(BKS:11.69)}
& \multicolumn{2}{c}{SC $\downarrow$~(BKS:123.30)} \\
\cmidrule(lr){2-3}\cmidrule(lr){4-5}\cmidrule(lr){6-7}\cmidrule(lr){8-9}
& Obj & Gap$_{abs}$ & Obj & Gap$_{abs}$ & Obj & Gap$_{abs}$ & Obj & Gap$_{abs}$ \\
\midrule

Gurobi (3600s)
& 98453.42 & 35.82
& 706.86 & 0.00
& 11.69 & 0.00
& 123.30 & 0.00 \\

Gurobi (1000s)
& 97407.51 & 1081.73
& 707.07 & 0.21
& 12.16 & 0.47
& 123.54 & 0.24 \\

\midrule
ND
& 94340.63 & 4148.61
& 707.12 & 0.26
& 14.18 & 2.49
& 123.51 & 0.21 \\

ND+SHSP
& \textbf{97121.05} & \textbf{1368.19}
& \textbf{707.07} & \textbf{0.21}
& \textbf{11.83} & \textbf{0.14}
& 123.51 & 0.21 \\
\midrule

Improvement
& - & 67.02\%
& - & 19.23\%
& - & 94.38\%
& - & 0.00\% \\
\midrule

PaS
& 97891.74 & 597.50
& 707.07 & 0.21
& 12.06 & 0.37
& 123.49 & 0.19 \\

PaS+SHSP
& \textbf{98487.72} & 1.52
& \textbf{706.99} & \textbf{0.13}
& \textbf{11.97} & 0.28
& \textbf{123.41} & 0.11\\
\midrule

Improvement
& - & 99.75\%
& - & 38.1\%
& - & 24.32\%
& - & 42.11\% \\
\midrule

Apollo
& 97997.74 & 491.50
& 707.07 & 0.21
& 11.97 & 0.28
& 123.42 & 0.12 \\

Apollo+SHSP
& \textbf{98489.24} & \textbf{0.00}
& \textbf{707.01} & \textbf{0.15}
& \textbf{11.71} & \textbf{0.02}
& \textbf{123.37 } & \textbf{0.07} \\
\midrule
Improvement
& - & 100.00\%
& - & 28.6\%
& - & 92.86\%
& - & 41.67\% \\

\bottomrule
\end{tabular}
\caption{Performance comparison on four benchmark
datasets (CA, WA, IP, and SC) using Gurobi under a 1000-second time limit. $\uparrow$ indicates that higher values are better, while
$\downarrow$ indicates that lower values are better.
Bold values denote the better results. 
Improvement denotes the relative reduction in Gap$_{\mathrm{abs}}$
relative to the corresponding downstream solution-prediction framework.}
\label{tab:main_results}
\end{table*}

\begin{figure*}[t]
	\centering
	\includegraphics[width=\textwidth]{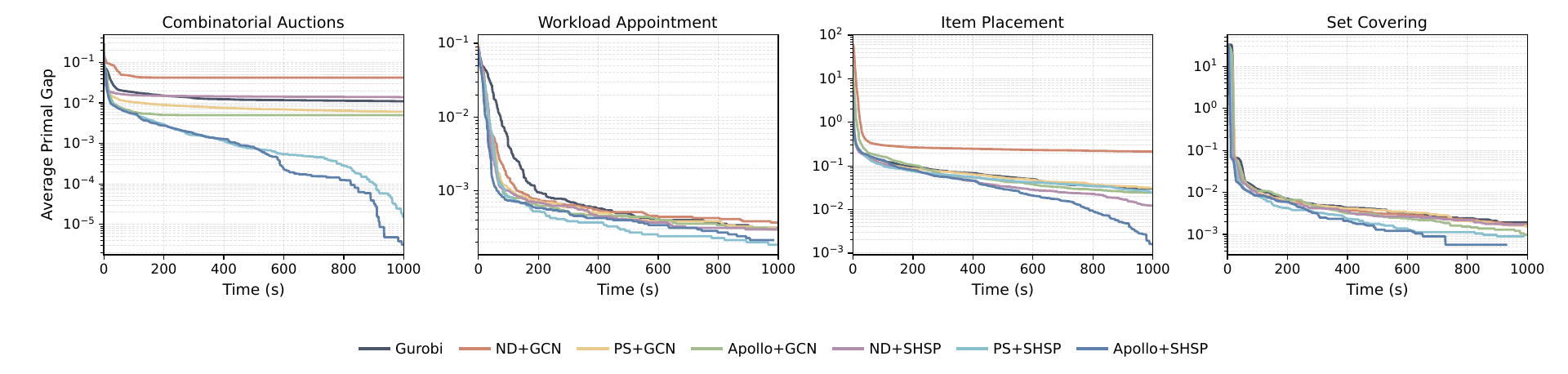}
    \caption{The convergence curves of primal gaps as the solving process proceeds.
All methods are implemented using Gurobi with a total time budget of
1,000 seconds. For a fair comparison, the graph construction time of SHSP
is included in this budget. Results are averaged over 100 testing
instances. A lower primal gap indicates faster convergence and better
solving performance.}
    \label{fig:primal_gap}
\end{figure*}

\begin{table*}[t]
\centering

\renewcommand{\arraystretch}{1.15}
\setlength{\tabcolsep}{4pt}

\small
\begin{tabular}{lcccccccc}
\toprule
\multirow{2}{*}{Method}
& \multicolumn{2}{c}{CA $\uparrow$~(BKS:98489.24)}
& \multicolumn{2}{c}{WA $\downarrow$~(BKS:706.86)}
& \multicolumn{2}{c}{IP $\downarrow$~(BKS:11.69)}
& \multicolumn{2}{c}{SC $\downarrow$~(BKS:123.30)} \\
\cmidrule(lr){2-3}\cmidrule(lr){4-5}\cmidrule(lr){6-7}\cmidrule(lr){8-9}
& Obj & Gap$_{abs}$ & Obj & Gap$_{abs}$ & Obj & Gap$_{abs}$ & Obj & Gap$_{abs}$ \\
\midrule

SCIP (3600s)
& 97795.78 & 693.46
& 708.21 & 1.35
& 19.06 & 7.37
& 124.58 & 1.28 \\

SCIP (1000s)
& 96932.37 & 1556.87
& 709.62 & 2.76
& 22.69 & 11.00
& 125.81 & 2.51 \\
\midrule

ND
& 94338.51 & 4150.73
& 707.83  & 0.97
& 16.33 & 4.64
& 125.87 & 2.57 \\

ND+SHSP
& \textbf{97267.46} & \textbf{1221.78}
& \textbf{707.18} & \textbf{0.32}
& \textbf{13.70} & \textbf{2.01}
& \textbf{125.84} & \textbf{2.54} \\
\midrule

Improvement
& - & 70.57\%
& - & 67.01\%
& - & 56.68\%
& - & 1.17\% \\
\midrule

PaS
& 97223.78 & 1265.46
& 708.27 & 1.41
& 19.61 & 7.92
& 125.82 & 2.52 \\

PaS+SHSP
& \textbf{97230.19} & \textbf{1259.05}
& \textbf{708.21} & \textbf{1.35}
& \textbf{19.44} & \textbf{7.75}
& \textbf{125.33} & \textbf{2.03} \\
\midrule

Improvement
& - & 0.51\%
& - & 4.26\%
& - & 2.15\%
& - & 19.44\% \\
\midrule

Apollo
& 96270.42 & 2218.82
& 708.42 & 1.56
& 20.48 & 8.79
& 125.74 & 2.44 \\

Apollo+SHSP
& \textbf{97192.31} & \textbf{1296.93}
& \textbf{708.21} & \textbf{1.35}
& \textbf{18.56} & \textbf{6.87}
& \textbf{125.45} & \textbf{2.15} \\
\midrule

Improvement
& - & 41.55\%
& - & 13.46\%
& - & 21.84\%
& - & 11.89\% \\

\bottomrule
\end{tabular}
\caption{Performance comparison on four benchmark
datasets (CA, WA, IP, and SC) using SCIP under a 1000-second time limit. $\uparrow$ indicates that higher values are better, while
$\downarrow$ indicates that lower values are better.
Bold values denote the better results. 
Improvement denotes the relative reduction in Gap$_{\mathrm{abs}}$
relative to the corresponding downstream solution-prediction framework.}
\label{tab:main_results_scip}
\end{table*}

\subsection{Main Evaluation}
\paragraph{Solving Performance}

To evaluate the effectiveness of our method, we replace the original GNN predictor and the variable fixing strategy in each framework with our SHSP predictor and structure-aware variable fixing strategy, forming \textbf{ND+SHSP}, \textbf{PaS+SHSP}, and \textbf{Apollo+SHSP}, respectively. We compare the solving performance between our framework and the baselines under a time limit of 1,000 seconds. 
Although SHSP performs hierarchical multi-step prediction, its additional neural network inference time is less than 0.2 seconds across all benchmark datasets.
The graph construction time ranges from 0.26 to 15.23 seconds and is included in the 1,000-second time budget for a fair comparison.
Table~\ref{tab:main_results} presents the average objective values and the corresponding average absolute primal gaps across four benchmark datasets, with Gurobi employed as the downstream solver. Similar results obtained with the SCIP solver are reported in Table~\ref{tab:main_results_scip}.
Detailed results of graph construction time and neural network inference time are provided in Appendix~\ref{subsec:Runtime Results}. 

Overall, SHSP consistently improves performance across all three learning-guided search frameworks compared with their original one-shot prediction baselines. With PaS, SHSP reduces the absolute primal gap by 99.7\% on CA, nearly closing the gap entirely, along with reductions of 38.1\%, 24.3\%, and 42.1\% on WA, IP, and SC, respectively. With Apollo, the gains are similarly substantial: the primal gap is completely eliminated on CA (100.0\% reduction) and reduced by 92.9\% on IP, with further reductions of 28.6\% and 41.7\% on WA and SC. When integrated with ND, SHSP also delivers solid improvements, reducing the absolute primal gap by 94.4\% on IP and 67.0\% on CA, and by 19.2\% on WA.
Remarkable improvements are also observed when using SCIP as the downstream solver. As shown in Table~\ref{tab:main_results_scip}, SHSP consistently reduces the absolute primal gap across all three frameworks on all four benchmarks, demonstrating that the proposed method generalizes well across different downstream solvers.
It is worth noting that Apollo+SHSP reaches the best known solution, surpassing the solution obtained by Gurobi with a 3,600-second time limit on the CA benchmark. 
This observation suggests that our method can provide a stronger initialization for the downstream solver, not only accelerating the search process but also guiding it toward higher-quality solutions.

\paragraph{Primal Gap as a Function of Runtime}
Figure~\ref{fig:primal_gap} presents the curves of the average primal gap, defined as Gap$_{\mathrm{rel}}
= \frac{\left|\mathrm{OBJ}-\mathrm{BKS}\right|}
{\left|\mathrm{BKS}\right|}$, throughout the solving process. Overall, SHSP-based methods consistently exhibit faster convergence and achieves lower primal gaps on the benchmarks than corresponding one-shot prediction methods. This behavior is attributed to more accurate solution predictions and more effective search-space reduction of our method.

\subsection{Ablation Study}
\paragraph{Confidence-Based Mask and Repair Mechanism}
To evaluate the effectiveness of the proposed mask-and-repair mechanism,
we compare three settings:
without mask and repair (None), with mask stage only
(Mask), and with both mask and repair
(Mask+Repair). As shown in Table~\ref{tab:mask_repair_main}, applying the masking mechanism alone does not consistently improve performance and may even cause slight degradation, since masked variables are simply discarded without correction, resulting in a loss of predictive information. In contrast, coupling masking with the proposed repair strategy consistently yields the best results across all datasets and both prediction frameworks, confirming that the two components are complementary and jointly contribute to the overall performance. More results are provided in Appendix~\ref{subsec:More Ablation Study Results}.

\begin{table}[t]
\centering
\renewcommand{\arraystretch}{1.15}
\setlength{\tabcolsep}{6pt}

\begin{tabular}{lcc}
\toprule
Method
& CA $\uparrow$
& IP $\downarrow$ \\
\midrule

PaS+SHSP (None)
& 98380.92
& 12.12 \\

PaS+SHSP (Mask)
& 98416.55
& 12.17 \\

PaS+SHSP (Mask+Repair)
& \textbf{98487.72}
& \textbf{11.97} \\

\midrule

Apollo+SHSP (None)
& 98385.20
& 11.78 \\

Apollo+SHSP (Mask)
& 98344.07
& 11.80 \\

Apollo+SHSP (Mask+Repair)
& \textbf{98489.24}
& \textbf{11.71} \\

\bottomrule
\end{tabular}
\caption{Ablation study of the proposed confidence-based
mask-and-repair strategy on the CA and IP datasets.
Bold values denote the better results.}
\label{tab:mask_repair_main}
\end{table}

\subsubsection{Hierarchical Solution Prediction Predictor}
To evaluate the effectiveness of the proposed hierarchical solution prediction predictor, we compare it with the original GNN predictor under the same structure-aware fixing strategy. As shown in Table~\ref{tab:ablation_main}, replacing original GNN predictor with hierarchical solution prediction predictor consistently improves the solution quality on both the PaS and Apollo frameworks. The improvement suggests that our predictor generates more accurate solution predictions, thereby enabling more effective variable fixing and subsequent problem reduction. Results on other problems are provided in Appendix~\ref{subsec:More Ablation Study Results}.

\begin{table}[t]
\centering
\setlength{\tabcolsep}{1.0mm}
\renewcommand{\arraystretch}{1.15}
\setlength{\tabcolsep}{6pt}

\begin{tabular}{lcc}
\toprule
Method (Fixing Strategy)
& CA $\uparrow$
& IP $\downarrow$ \\
\midrule

PaS (Structure-Aware)
& 97184.33
& 12.12 \\

PaS+SHSP (Original)
& 98061.51
& 12.34 \\

PaS+SHSP (Structure-Aware)
& \textbf{98487.72}
& \textbf{11.97} \\

\midrule

Apollo (Structure-Aware)
& 98015.32
& 11.87 \\

Apollo+SHSP (Original)
& 98070.43
& 12.08 \\

Apollo+SHSP (Structure-Aware)
& \textbf{98489.24}
& \textbf{11.71} \\

\bottomrule
\end{tabular}
\caption{Ablation study of the proposed hierarchical solution prediction and structure-aware fixing strategy on the CA and IP datasets.
Bold values denote the better results.}
\label{tab:ablation_main}
\end{table}

\paragraph{Structure-Aware Fixing Strategies}
To evaluate the effectiveness of the proposed structure-aware fixing strategy, we compare it with the original fixing strategy while using the same predictor. As shown in Table~\ref{tab:ablation_main}, the proposed structure-aware fixing strategy consistently outperforms the normal fixing strategy on both the PaS and Apollo frameworks. The improvement suggests that prioritizing structurally important variables leads to more effective search space reduction, allowing the solver to explore the remaining search space more efficiently and ultimately achieve better solving performance. Results on other problems are provided in Appendix~\ref{subsec:More Ablation Study Results}.

\section{Conclusion}
In this paper, we propose SHSP, a structure-aware hierarchical solution prediction framework for MILP solving. SHSP replaces the parallel marginal decoding with a hierarchical conditional decoding mechanism guided by a variable coupling graph, decoding variables from weakly to strongly coupled ones while conditioning each step on previously predicted assignments. It further incorporates a mask-and-repair mechanism to mitigate error accumulation, and a structure-aware fixing strategy to enable more effective search-space reduction in downstream frameworks. Extensive experiments on four standard benchmarks show that SHSP consistently outperforms one-shot baselines across three frameworks and two downstream solvers, achieving a 54\% average reduction in the absolute primal gap with negligible inference overhead. In future work, we plan to learn coupling scores in an end-to-end manner and extend to broader problem classes.

\appendix
\bibliography{aaai2027}
\newpage
\section{Related Work}
\label{sec:Related Works}
\subsection{Learning alongside MILP Solvers}
In recent years, machine learning has emerged as a promising paradigm for accelerating MILP solving~\cite{li2024optverse}, and existing studies generally fall into two categories. The first category learns heuristic policies for key components of branch-and-bound (B\&B) solvers,
including branching variable selection
\cite{khalil2016learning,gasse2019exact,gupta2020hybrid,zarpellon2021parameterizing,gupta2022lookback,
scavuzzo2022tree,lin2024cambranch,
zhang2024towards,kuang2024rethinking},
cutting plane selection
\cite{tang2020reinforcement,wang2023learning,
wang2024learning,huang2022learning,
paulus2022learning,ling2024learning,
puigdemont2024learning}, scheduling of primal heuristics
\cite{khalil2017learning,chmiela2021learning}. These methods typically leverage imitation learning or reinforcement learning to learn effective heuristic policies from historical solving trajectories or expert policies. 

Among all components, branching and cut selection have received the most attention due to their significant impact on solver performance. For branching, \citet{khalil2016learning} pioneered the imitation of strong branching decisions via a learned ranking function, and more recently, \citet{hou2026llm4branch} leveraged large language models to automatically discover branching policies. For cut selection, \citet{tang2020reinforcement} proposed to learn adaptive selection policies through reinforcement learning, while \citet{miracle2026} further combined imitation learning and reinforcement learning to obtain memory-efficient policies. 
Collectively, these studies demonstrate that learning effective policies for individual solver components can substantially enhance the performance of modern MILP solvers.

\subsection{End-to-End Prediction for MILP}
The second category employs machine learning models such as graph neural networks (GNNs) to directly predict high-quality solutions for MILP instances~\cite{ding2020accelerating,zeng2024effective,geng2025differentiable}. Research in this category advances along two complementary aspects. The first aspect concerns how predictions are exploited to reduce the original problem~\cite{nair2020solving,han2023gnn,liu2025apollo}. \citet{nair2020solving} proposed to directly fix high-confidence variables to their predicted values, yielding a reduced sub-problem, which was subsequently refined in follow-up studies~\cite{yoon2021confidence,paulus2023learning}.
\citet{han2023gnn} and \citet{huang2024contrastive} relaxed such hard fixing by searching within a trust region centered at the predicted partial assignment, and \citet{liu2025apollo} further extended this paradigm by alternating between prediction and solver-based correction.

The second aspect focuses on improving the prediction capacity~\cite{pu2026coco,wang2026milpnet}. \citet{pu2026coco} enhanced prediction quality by explicitly modeling inter-variable correlations, while \citet{wang2026milpnet} replaced the conventional bipartite graph representation with geometric feature sequences and performed prediction via a multi-scale attention architecture. Together, these efforts have established learning-based solution prediction as an effective paradigm for accelerating MILP solving.

\section{Graph Representation Details}
\label{sec:Graph Representation Details}

The bipartite graph representation used in this paper
is based on the representation adopted by Predict-and-Search~\cite{han2023gnn}. To support hierarchical decoding, we further introduce three additional variable features: target indicator, previous prediction,
and previous confidence. We list the graph features in Table~\ref{tab:Graph features}.

\begin{table}[htbp]
\centering

\renewcommand{\arraystretch}{1.15}
\setlength{\tabcolsep}{6pt}
\resizebox{\linewidth}{!}{
\begin{tabular}{clp{3.2cm}p{6.5cm}}
\toprule
Index & \multicolumn{1}{c}{Variable Feature Name} &
\multicolumn{2}{c}{Description} \\
\midrule
0    & Objective                    & \multicolumn{2}{l}{Normalized objective coefficient} \\
1    & Variable coefficient         & \multicolumn{2}{l}{Average variable coefficient in all constraints} \\
2    & Variable degree              & \multicolumn{2}{p{6.5cm}}{Degree of the variable node in the bipartite graph representation} \\
3    & Maximum variable coefficient & \multicolumn{2}{l}{Maximum variable coefficient in all constraints} \\
4    & Minimum variable coefficient & \multicolumn{2}{l}{Minimum variable coefficient in all constraints} \\
5    & Variable type                & \multicolumn{2}{l}{Whether the variable is an integer variable or not} \\
\textbf{6}    & \textbf{Target indicator}             & \multicolumn{2}{p{6.5cm}}{\textbf{Whether the variable belongs to the current prediction hierarchy}} \\
\textbf{7}   & \textbf{Previous prediction}          & \multicolumn{2}{p{6.5cm}}{\textbf{Previously predicted value of the variable}} \\
\textbf{8}    & \textbf{Previous confidence}          & \multicolumn{2}{p{6.5cm}}{\textbf{Confidence score of the predicted value}} \\
9--24 & Position embedding          & \multicolumn{2}{p{6.5cm}}{Binary encoding of the order of appearance for each variable among all variables} \\
\midrule
Index &
\multicolumn{1}{c}{Constraint Feature Name} &
\multicolumn{2}{c}{Description} \\
\midrule
0 & Constraint coefficient & \multicolumn{2}{l}{Average of all coefficients in the constraint} \\
1 & Constraint degree      & \multicolumn{2}{l}{Degree of constraint nodes} \\
2 & Bias                   & \multicolumn{2}{l}{Normalized right-hand side of the constraint} \\
3 & Sense                  & \multicolumn{2}{l}{The sense of the constraint} \\
\midrule
Index &
\multicolumn{1}{c}{Edge Feature Name} &
\multicolumn{2}{c}{Description} \\
\midrule
0 & Coefficient & \multicolumn{2}{l}{Constraint coefficient} \\
\bottomrule
\end{tabular}}
\caption{The variable features, constraint features, and edge features used for the graph representation. Newly introduced features are shown in bold.}
\label{tab:Graph features}
\end{table}

\section{Variable Coupling Graph Details}
\label{sec:Variable Coupling Graph Details}

\paragraph{Expected Violation}
The \emph{expected violation} measures how strongly a variable pair is restricted by a constraint. Violation of the variable pair $(z_i,z_j)$ in constraint $c$ is defined as
\begin{equation*}
\Delta
\left(
\phi_{ij}^{(c)}
\right)
=
\begin{cases}
\max(b_c-\phi_{ij}^{(c)},0),
& \circ_c\text{ is }\geq,\\[1mm]
\max(\phi_{ij}^{(c)}-b_c,0),
& \circ_c\text{ is }\leq,\\[1mm]
\left|\phi_{ij}^{(c)}-b_c\right|,
& \circ_c\text{ is}= .
\end{cases}
\end{equation*}
where $\phi_{ij}^{(c)} =a_{ci}z_i+a_{cj}z_j$,
with $a_{ci}$ and $a_{cj}$ denoting the coefficients of
variables $z_i$ and $z_j$ in constraint $c$, respectively.

The expected violation
$P_{ij}^{(c)}
=
\mathbb{E}
\left[
\Delta
\left(
\phi_{ij}^{(c)}
\right)
\right]$ is computed as follows. 
For a binary--binary pair $(z_i,z_j)$, we assume
$z_i,z_j\sim\mathrm{Uniform}(\{0,1\})$, and the expected violation is computed as

\begin{equation*}
P_{ij}^{(c)}
=
\frac{1}{4}
\sum_{z_i,z_j\in\{0,1\}}
\Delta
\left(
\phi_{ij}^{(c)}
\right).
\end{equation*}

For a binary--continuous pair $(z_i,z_j)$, we assume
$z_i\sim\mathrm{Uniform}(\{0,1\})$ and
$z_j\sim\mathrm{Uniform}(l_j,u_j)$, where $l_j$ and $u_j$ denote the lower and upper bounds of $z_j$. The expected violation is computed as

\begin{equation*}
P_{ij}^{(c)}
=
\frac{1}{2}
\sum_{z_i\in\{0,1\}}
\frac{1}{u_j-l_j}
\int_{l_j}^{u_j}
\Delta
\left(
\phi_{ij}^{(c)}
\right)
\,dz_j .
\end{equation*}

To eliminate scale differences across constraints and
measure the relative strength of a variable pair within the constraint, we normalize the expected violation as

\begin{equation*}
\widehat{P}_{ij}^{(c)}
=
\frac{
P_{ij}^{(c)}
}{
\frac{1}{|E_c|}
\sum\limits_{(u,v)\in E_c}
P_{uv}^{(c)}
},
\end{equation*}
where $E_c$ denotes the set of retained variable pairs in constraint $c$.

\paragraph{Coefficient Importance}
The \emph{coefficient importance} characterizes the
relative contribution of a variable pair within a constraint. For each variable $z_i$, we define a scale factor $s_i$ according to its variable type, where $s_i$ is set to $1$ for binary variables and to $u_i-l_i$ for continuous variables.
Then we compute the coefficient importance $r_{ij}^{(c)}$ of the variable pair and normalize it within the constraint to obtain $\widehat r_{ij}^{(c)}$, which represents the relative importance of a variable pair within the constraint:

\begin{equation*}
r_{ij}^{(c)}
=
\frac{
|a_{ci}|s_i+|a_{cj}|s_j
}{2},
\end{equation*}
\begin{equation*}
\widehat r_{ij}^{(c)}
=
\frac{
r_{ij}^{(c)}
}{
\frac{1}{|E_c|}
\sum\limits_{(u,v)\in E_c}
r_{uv}^{(c)}
}.
\end{equation*}

\paragraph{Edge Weight Aggregation.}
Finally, the local coupling weight of a pair in constraint $c$ is computed as $w_{ij}^{(c)}
=\widehat{P}_{ij}^{(c)}
\widehat r_{ij}^{(c)}$, and the final edge weight $W_{ij}$ between variables $z_i$ and $z_j$ is obtained by summing the local weights over all constraints containing both variables.

\section{Benchmark Details}
\label{sec:Benchmark Details}

\subsection{Benchmarks in Main Evaluation}
\label{subsec:Benchmarks in Main Evaluation}
The CA and SC benchmark instances are generated following the process described by~\citet{gasse2019exact}. Specifically, the CA dataset follows the generation method of~\citet{leytonbrown2000towards}, while the SC dataset is constructed using the algorithm presented by~\citet{balas1980set}. The IP and WA instances are obtained from the NeurIPS ML4CO 2021 competition~\cite{gasse2022ml4co}. The statistical information is provided in Table~\ref{tab:benchmarks statistics}.

\begin{table}[htbp]
\centering
\renewcommand{\arraystretch}{1.1}
\setlength{\tabcolsep}{6pt}
\begin{tabular}{lcccc}
\toprule
 & CA & WA & IP & SC \\
\midrule
Constraint Number    & 2592  & 64318 & 195  & 3000 \\
Variable Number      & 1500  & 61000 & 1083 & 5000 \\
Binary Variables     & 1500  & 1000  & 1050 & 5000 \\
Continuous Variables & 0     & 60000 & 33   & 0 \\
Integer Variables    & 0     & 0     & 0    & 0 \\
\bottomrule
\end{tabular}
\caption{Summary statistics of the benchmark datasets used in the main evaluation.}
\label{tab:benchmarks statistics}
\end{table}

\subsection{Subset of MIPLIB}
\label{subsec:Subset of MIPLIB}
We use the IIS dataset, which is a subset of MIPLIB~\cite{gleixner2021miplib} constructed by~\citet{liu2025apollo}, to evaluate the solvers’ ability to handle challenging real-world instances. According to~\citet{liu2025apollo}, the instances in the dataset are selected based on their similarity, which is measured using the 100 human-designed features~\cite{gleixner2021miplib}. Instances with presolving times exceeding 300 seconds or those that exceed GPU memory limits during the inference process are discarded. The IIS dataset contains eleven instances, including eight training instances and three testing instances (ramos3,
scpj4scip, and scpl4).
Detailed statistics of the IIS dataset are reported in Table~\ref{tab:iis_statistics}.

\begin{table}[htbp]
\centering

\renewcommand{\arraystretch}{1.1}
\setlength{\tabcolsep}{5pt}

\resizebox{\linewidth}{!}{
\begin{tabular}{lrrrrr}
\toprule
Instance &
Constraint Number &
Variable Number &
Binary Variables \\
\midrule
ex1010-pi       & 1468 & 25200  & 25200   \\
fast0507        & 507  & 63009  & 63009   \\
glass-sc        & 6119 & 214    & 214    \\
iis-glass-cov   & 5375 & 214    & 214    \\
iis-hc-cov      & 9727 & 297    & 297    \\
ramos3          & 2187 & 2187   & 2187  \\
scpj4scip       & 1000 & 99947  & 99947   \\
scpk4           & 2000 & 100000 & 100000 \\
scpl4           & 2000 & 200000 & 200000 \\
seymour         & 4944 & 1372   & 1372   \\
v150d30-2hopcds & 7822 & 150    & 150    \\
\bottomrule
\end{tabular}
}

\caption{Statistical information of the instances in the IIS dataset. All instances contain only binary variables.}
\label{tab:iis_statistics}
\end{table}

\section{Implementation Details}
\label{sec:Implementation Details}
\subsection{SHSP Predictor Details}
\label{subsec:SHSP Predictor Details}
The predictor used in SHSP is based on the graph neural network architecture adopted in previous learning-based MILP approaches~\cite{han2023gnn,liu2025apollo}.
To support the hierarchical decoding process in SHSP, we introduce a learnable step embedding, which enables the network to adapt its prediction behavior across different decoding steps. Specifically, the initial representation of each variable node is computed as
\[
h_{v_i}^{(0,k)}
=
f_s\left(
\left[
f_v\left(x_i\right);
\mathrm{Emb}(k)
\right]
\right).
\]
Here, $v_i$ denotes the $i$-th variable node, $k$ is the current decoding step, and $x_i$ is the feature vector of $v_i$. The function $f_v(\cdot)$ denotes the variable embedding network, $\mathrm{Emb}(k)$ is the learnable embedding corresponding to decoding step $k$, and $f_s(\cdot)$ fuses the variable and step embeddings.
The resulting representation $h_{v_i}^{(0,k)}$ is then
used as the input to the same bipartite message-passing
architecture adopted in previous
approaches~\cite{han2023gnn,liu2025apollo}.

\subsection{Training Details}
\label{subsec:Training Details}
During predictor training, we set the initial learning rate to be 0.001 and the number of training epochs to 500. To collect the training data, we run a single-threaded Gurobi~\cite{gurobi2021}
on each training and validation instance for 3,600 seconds and record the best 50 solutions.

For SHSP training, we incorporate a \textit{teacher forcing} strategy that employs ground-truth assignments as conditioning information in the early stage of training, and the teacher forcing ratio $\rho_t$ is progressively decayed as training proceeds. Specifically, with probability
$\rho_t$, we use the ground-truth value as the history input for the next step;
otherwise, we use the model prediction. The teacher forcing ratio
$\rho_t$ is linearly decayed from $\rho_{\mathrm{start}}$ to
$\rho_{\mathrm{end}}$ during the first $T_{\mathrm{decay}}$ epochs:
\[
\rho_t =
\rho_{\mathrm{start}}
+
\min\left(1, \frac{t}{T_{\mathrm{decay}}}\right)
\left(
\rho_{\mathrm{end}} - \rho_{\mathrm{start}}
\right),
\]
where \(t\) is the current epoch. During validation
and inference, we disable teacher forcing and always use the model predictions
to update the conditioning context. In experiments, we set
$\rho_{\mathrm{start}}=1.0$,
$\rho_{\mathrm{end}}=0$,
and
$T_{\mathrm{decay}}=50$.

\subsection{Integration with Downstream Learning-based Search Methods}
\label{subsec:Integrate with Downstream Learning-based Search Methods}

We replace the original GNN predictor and the variable fixing strategy in each framework with our SHSP predictor and structure-aware variable fixing strategy, forming \textbf{ND+SHSP}, \textbf{PaS+SHSP}, and \textbf{Apollo+SHSP}, respectively.

For ND+SHSP and PaS+SHSP, the variable coupling graph is constructed once before neural network inference for each instance. The graph construction time is included in the solver time budget.

For Apollo+SHSP, the variable coupling graph is
constructed before the first prediction and is subsequently updated for each reduced subproblem before prediction. Similarly,
the time required for graph construction and updates is
deducted from the corresponding solver time budget.

\subsection{Inference Details}
\label{subsec:Inference Details}

We conducted all the experiments on a single
machine with an AMD EPYC 7513 32-Core Processor and NVIDIA GeForce RTX 4090 GPUs.
The inference settings are described as follows.

For the baselines, the hyperparameters $(k_0,k_1)$ for ND and $(k_0,k_1,\Delta)$ for PaS are summarized in
Table~\ref{tab:nd_ps_hyperparameters}. For Apollo-MILP, the number of iterations is set to 4.
The hyperparameter settings
$(k_0^{(i)}, k_1^{(i)}, \Delta^{(i)})$
for each iteration are summarized in
Table~\ref{tab:apollo_hyperparameters}.

\begin{table}[htbp]
\centering

\renewcommand{\arraystretch}{1.15}
\setlength{\tabcolsep}{10pt}

\begin{tabular}{lcc}
\toprule
Dataset & PaS $(k_0,k_1,\Delta)$ & ND $(k_0,k_1)$ \\
\midrule
CA & $(600,0,20)$ & $(600,0)$ \\
WA & $(0,500,10)$ & $(0,500)$ \\
IP & $(400,5,10)$ & $(400,5)$ \\
SC & $(2000,0,100)$ & $(2000,0)$ \\
\bottomrule
\end{tabular}

\caption{Hyperparameter settings for PaS and ND on different benchmark datasets.}
\label{tab:nd_ps_hyperparameters}
\end{table}

\begin{table}[htbp]
\centering

\renewcommand{\arraystretch}{1.15}
\setlength{\tabcolsep}{8pt}
\resizebox{\linewidth}{!}{%
\begin{tabular}{lcccc}
\toprule
 & CA & WA & IP & SC \\
\midrule
Iteration 1 &
$(400,0,60)$ &
$(20,200,100)$ &
$(100,20,50)$ &
$(1000,0,200)$ \\

Iteration 2 &
$(200,0,30)$ &
$(10,100,50)$ &
$(40,15,20)$ &
$(500,0,100)$ \\

Iteration 3 &
$(100,0,15)$ &
$(10,5,5)$ &
$(20,15,10)$ &
$(250,0,50)$ \\

Iteration 4 &
$(50,0,10)$ &
$(1,10,5)$ &
$(5,50,30)$ &
$(10,0,5)$ \\
\bottomrule
\end{tabular}}
\caption{Hyperparameter configuration of $(k_0^{(i)}, k_1^{(i)}, \Delta^{(i)})$ in Apollo-MILP.}
\label{tab:apollo_hyperparameters}
\end{table}

For our method, we use the same hyperparameter settings as the corresponding baseline framework. The number of decoding steps is set to 2 for all benchmark datasets except WA, for which it is set to 4. The additional hyperparameter settings $(\theta_0,\theta_1,\eta)$ are reported in Table~\ref{tab:theta_0,theta_1,eta}, where $\eta$ controls the VCS threshold. Specifically, let \(z_{(1)}, \ldots, z_{(n)}\) denote the variables sorted in descending order of VCS. We use the VCS of \(z_{(\lceil \eta n \rceil)}\) as the threshold.

\begin{table}[htbp]
\centering

\renewcommand{\arraystretch}{1.15}
\setlength{\tabcolsep}{10pt}
\resizebox{\linewidth}{!}{%
\begin{tabular}{lccc}
\toprule
Dataset & ND+SHSP & PaS+SHSP & Apollo+SHSP \\
\midrule
CA & $(0.10,0.90,0.5)$ & $(0.10,0.90,0.5)$ & $(0.04,0.90,0.7)$ \\
WA & $(0.10,0.90,0.7)$ & $(0.10,0.90,0.7)$ & $(0.10,0.95,0.8)$ \\
IP & $(0.05,0.95,0.7)$ & $(0.10,0.90,0.7)$ & $(0.10,0.50,0.5)$ \\
SC & $(0.10,0.90,0.5)$ & $(0.10,0.90,0.5)$ & $(0.10,0.90,0.7)$ \\
\bottomrule
\end{tabular}}

\caption{The additional hyperparameter settings $(\theta_0,\theta_1,\eta)$ for ND+SHSP, PaS+SHSP  and Apollo+SHSP on different benchmark datasets.}
\label{tab:theta_0,theta_1,eta}
\end{table}

\vspace{-0.5em}

\section{Additional Experimental Results}
\label{sec:Additional Experimental Results}
\subsection{More Ablation Study Results}
\label{subsec:More Ablation Study Results}
\begin{table*}[ht]
\centering

\renewcommand{\arraystretch}{1.15}
\setlength{\tabcolsep}{4pt}
\small
\begin{tabular}{lcccccccc}
\toprule
\multirow{2}{*}{Method}
& \multicolumn{2}{c}{CA $\uparrow$(BKS:98489.24)}
& \multicolumn{2}{c}{WA $\downarrow$(BKS:706.86)}
& \multicolumn{2}{c}{IP $\downarrow$(BKS:11.69)}
& \multicolumn{2}{c}{SC $\downarrow$(BKS:123.30)} \\
\cmidrule(lr){2-3}\cmidrule(lr){4-5}\cmidrule(lr){6-7}\cmidrule(lr){8-9}
& Obj & Gap$_{abs}$ & Obj & Gap$_{abs}$ & Obj & Gap$_{abs}$ & Obj & Gap$_{abs}$ \\
\midrule

PaS+SHSP (None)
& 98380.92 & 108.32
& 707.06 & 0.20
& 12.12 & 0.43
& 123.47 & 0.17 \\

PaS+SHSP (Mask)
& 98416.55 & 72.69
& 707.05 & 0.19
& 12.17 & 0.48
& 123.53 & 0.23 \\

PaS+SHSP (Mask+Repair)
& \textbf{98487.72} & \textbf{1.52}
& \textbf{706.99} & \textbf{0.13}
& \textbf{11.97} & \textbf{0.28}
& \textbf{123.41} & \textbf{0.11} \\

\midrule

Apollo+SHSP (None)
& 98385.20 & 104.04
& 707.10 & 0.24
& 11.78 & 0.09
& 123.42 & 0.12 \\

Apollo+SHSP (Mask)
& 98344.07 & 145.17
& \textbf{707.01} & \textbf{0.15}
& 11.80 & 0.11
& 123.39 & 0.09 \\

Apollo+SHSP (Mask+Repair)
& \textbf{98489.24} & \textbf{0.00}
& \textbf{707.01} & \textbf{0.15}
& \textbf{11.71} & \textbf{0.02}
& \textbf{123.37} & \textbf{0.07} \\

\bottomrule
\end{tabular}
\caption{Detailed ablation study of the proposed confidence-based mask-and-repair strategy. $\uparrow$ indicates that higher values are better, while
$\downarrow$ indicates that lower values are better.
Bold values denote the best results, including ties.}
\label{tab:mask_repair}

\end{table*}

\begin{table*}[ht]
\centering

\renewcommand{\arraystretch}{1.15}
\setlength{\tabcolsep}{4pt}
\small
\begin{tabular}{lcccccccc}
\toprule
\multirow{2}{*}{Method}
& \multicolumn{2}{c}{CA $\uparrow$(BKS:98489.24)}
& \multicolumn{2}{c}{WA $\downarrow$(BKS:706.86)}
& \multicolumn{2}{c}{IP $\downarrow$(BKS:11.69)}
& \multicolumn{2}{c}{SC $\downarrow$(BKS:123.30)} \\
\cmidrule(lr){2-3}\cmidrule(lr){4-5}\cmidrule(lr){6-7}\cmidrule(lr){8-9}
& Obj & Gap$_{abs}$ & Obj & Gap$_{abs}$ & Obj & Gap$_{abs}$ & Obj & Gap$_{abs}$ \\
\midrule

PaS (structure-aware fixing)
& 97184.33 & 1304.91
& 707.08 & 0.22
& 12.12 & 0.43
& 123.49 & 0.19 \\

PaS+SHSP (original fixing)
& 98061.51 & 427.73
& 707.01 & 0.15
& 12.34 & 0.65
& 123.53 & 0.23 \\

PaS+SHSP (structure-aware fixing)
& \textbf{98487.72} & \textbf{1.52}
& \textbf{706.99} & \textbf{0.13}
& \textbf{11.97} & \textbf{0.28}
& \textbf{123.41} & \textbf{0.11} \\

\midrule

Apollo (structure-aware fixing)
& 98015.32 & 473.92
& 707.09 & 0.23
& 11.87 & 0.18
& 123.40 & 0.10 \\

Apollo+SHSP (original fixing)
& 98070.43 & 418.81
& \textbf{707.01} & \textbf{0.15}
& 12.08 & 0.39
& \textbf{123.37} & \textbf{0.07} \\

Apollo+SHSP (structure-aware fixing)
& \textbf{98489.24} & \textbf{0.00}
& \textbf{707.01} & \textbf{0.15}
& \textbf{11.71} & \textbf{0.02}
& \textbf{123.37} & \textbf{0.07} \\

\bottomrule
\end{tabular}
\caption{Detailed ablation study of the proposed hierarchical solution prediction predictor and structure-aware fixing strategy. $\uparrow$ indicates that higher values are better, while
$\downarrow$ indicates that lower values are better.
Bold values denote the best results, including ties.}
\label{tab:ablation}
\end{table*}

\paragraph{Confidence-Based Mask and Repair Mechanism }
We perform an ablation study to evaluate the proposed mask-and-repair mechanism. Specifically, we compare three variants: None (without mask or repair), Mask (mask only), and Mask+Repair (mask and repair) on the four benchmark datasets (CA, WA, IP, and SC). The results are shown in Table~\ref{tab:mask_repair}. The observations are consistent with those discussed in the main paper: the two components are complementary and jointly contribute to the overall performance.

\paragraph{Hierarchical Solution Prediction Predictor}
To evaluate the effectiveness of the proposed hierarchical solution prediction predictor, we compare it with the original GNN predictor under the same structure-aware fixing strategy on the four benchmark datasets (CA, WA, IP, and SC). The results are shown in Table~\ref{tab:ablation}, which suggests that our predictor generates more accurate solution predictions, thereby enabling more effective
variable fixing and problem reduction.

\paragraph{Fixing Strategies}
To evaluate the effectiveness of the proposed structure-aware fixing strategy, we compare it with the original fixing strategy while using the same predictor. The results on the four benchmark datasets (CA, WA, IP, and SC) are shown in Table~\ref{tab:ablation}. The observations are consistent with those discussed in the
main paper: prioritizing structurally important variables leads to more effective search space reduction, allowing the solver to explore the remaining search space more efficiently and achieve better performance.

\subsection{Results on MIPLIB}
To further demonstrate the applicability of our method, we conduct experiments on instances from challenging real-world dataset
MIPLIB~\cite{gleixner2021miplib}. However, applying ML-based solvers directly to the entire dataset can be difficult because of the heterogeneous nature
of the instances in MIPLIB. Therefore, following~\citet{liu2025apollo}, we focus on a subset of MIPLIB that contains similar instances. More information on the selected MILP subset, referred to as IIS, is provided in Appendix~\ref{subsec:Subset of MIPLIB}. We report the solving performance of the solvers in Table~\ref{tab:iis_results} and Table~\ref{tab:detailed_iis_results}, where our methods
outperform their corresponding baselines, showcasing the potential for real-world applications.

\begin{table}[htbp]
\centering

\renewcommand{\arraystretch}{1.20}
\setlength{\tabcolsep}{10pt}
\small
\resizebox{0.7\linewidth}{!}{%
\begin{tabular}{lcc}
\toprule
Method & Obj$\downarrow$ & Gap$_{\mathrm{abs}}\downarrow$ \\
\midrule
Gurobi & 211.00 & 20.00 \\
PaS     & 209.00 & 18.00 \\
PaS+SHSP     & \textbf{206.67} & \textbf{15.67} \\
\midrule
Apollo  & 214.33 & 23.33 \\
Apollo+SHSP  & \textbf{208.67} & \textbf{17.67} \\
\bottomrule
\end{tabular}}
\caption{
Results on the IIS dataset, a subset of MIPLIB used
by~\citet{liu2025apollo}. All learning-based methods use Gurobi as the downstream solver, with a solving time limit of 3,600 seconds.
}
\label{tab:iis_results}
\end{table}

\begin{table}[htbp]
\centering
\resizebox{\linewidth}{!}{%
\begin{tabular}{lcccccc}
\toprule
Instance & BKS & Gurobi & PaS & Apollo & PaS+SHSP & Apollo+SHSP\\
\midrule
ramos3    & 186.00 & 228.00 &  227.00 & 242.00 & \textbf{221.00} & 226.00\\
scpj4scip & 128.00 & 133.00 &  131.00 & 132.00 & \textbf{130.00} & 131.00\\
scpl4     & 259.00 & 272.00 &  269.00 & 269.00 & 269.00 & 269.00\\
\bottomrule
\end{tabular}}
\caption{The best objectives found on each test instance in IIS. BKS denotes the best known objective values reported by MIPLIB (\url{https://miplib.zib.de/index.html}).}
\label{tab:detailed_iis_results}
\end{table}
\vspace{-1.5em}
\subsection{Runtime Results}
\label{subsec:Runtime Results}
We report network inference time and graph construction time in Table~\ref {tab:predict_time}. The inference time of the SHSP predictor is only slightly higher than that of the original GNN predictor, which remains negligible compared with the overall solver runtime. In addition, the graph construction time ranges from 0.26s to 15.23s per instance, which is also much shorter than the solver runtime.

\begin{table}[htbp]
\centering

\renewcommand{\arraystretch}{1.15}
\setlength{\tabcolsep}{10pt}
\small
\resizebox{\linewidth}{!}{
\begin{tabular}{lccc}
\toprule
Dataset & GNN Predictor & SHSP Predictor & Graph construction time\\
\midrule
CA & 0.0050s & 0.0146s & 0.82s\\
WA & 0.0439s & 0.2197s & 5.48s\\
IP & 0.0043s & 0.0131s & 0.26s\\
SC & 0.0658s & 0.1953s & 15.23s\\
\bottomrule
\end{tabular}}

\caption{Comparison of the average per-instance inference time and graph construction time (in seconds) on the four benchmark datasets.}
\label{tab:predict_time}

\end{table}

\end{document}